\documentclass[11pt,a4paper]{article}
\usepackage[hyperref]{acl2019}
\usepackage{times}
\usepackage{amsmath}
\usepackage{amssymb}
\usepackage{latexsym}
\usepackage[noend]{algpseudocode}
\usepackage{algorithm,algorithmicx}
\usepackage{graphicx, caption, subcaption}
\usepackage{booktabs}
\usepackage{url}
\usepackage{linearb}
\usepackage{verbatim}
\usepackage{pifont}

\usepackage{xpatch}
\makeatletter
\xpatchcmd{\algorithmic}{\itemsep\z@}{\itemsep=1.5pt}{}{}
\makeatother

\newcommand*\Let[2]{\State #1 $\gets$ #2}
\DeclareMathOperator{\E}{\mathbb{E}}
\DeclareMathOperator*{\argmax}{arg\,max}

\usepackage{bbm}

\usepackage{mdframed}
\usepackage{lipsum}

\let\oldeqref\eqref
\renewcommand{\eqref}[1]{Equation~\oldeqref{#1}}
\newcommand{\tabref}[1]{Table~\ref{#1}}
\newcommand{\figref}[1]{Figure~\ref{#1}}

\newcommand{\mF}{\mathcal{F}}
\newcommand{\mX}{\mathcal{X}}
\newcommand{\mY}{\mathcal{Y}}
\newcommand{\card}[1]{\lvert#1\rvert}
\newcommand{\norm}[1]{\|#1\|_2}

\newcommand{\rv}[1]{{\color{black}#1}}

\aclfinalcopy % Uncomment this line for the final submission
\title{Neural Decipherment via Minimum-Cost Flow: from Ugaritic to Linear B}

\author{Jiaming Luo \\
  CSAIL, MIT \\
  \texttt{j\_luo@csail.mit.edu} \\\And
  Yuan Cao \\
  Google Brain \\
  \texttt{yuancao@google.com} \\\And
  Regina Barzilay \\
  CSAIL, MIT \\
  \texttt{regina@csail.mit.edu}\\}
\date{}

\begin{document}
\maketitle

% Q:
% - What's the motivation? Applying neural models for decipherment?
% - Scope: only for cognate identification or general lexicon induction? Not sure if only identifying cognates would be enough for comparison

\begin{abstract}

In this paper we propose a novel neural approach for automatic decipherment of lost languages. To compensate for the lack of strong supervision signal, our model design is informed by patterns in language change documented in historical linguistics.  The model utilizes an expressive sequence-to-sequence model to capture character-level correspondences between cognates. To effectively train the model in an unsupervised manner, we innovate the training procedure by formalizing it as a minimum-cost flow problem.
When applied to the decipherment of Ugaritic, we achieve a 5.5\% absolute improvement over state-of-the-art results. We also report the first automatic results in deciphering Linear B, a syllabic language related to ancient Greek, where our model correctly translates  67.3\% of cognates.\footnote{Code and all datasets are hosted in \url{https://github.com/j-luo93/NeuroDecipher}.}

\begin{comment}
We introduce a neural decipherment approach. Our approach is language-independent and built upon fundamental principles of decipherment across multiple languages, which effectively guide the model decipherment process without supervision. We employ a neural sequence-to-sequence model to capture character-level cognate generation process, and innovate a training procedure is formulated as flow to impose lexicon-level structural sparsity. We demonstrate the efficacy of our approach on two lost languages -- Ugaritic  and Linear B, from different linguistic families, and observed substantially high accuracy in cognate identification.

% We propose a novel approach for language decipherment. Our approach uses an expressive neural sequence-to-sequence model to capture character-level correspondence between the cipher and known languages. In order to effectively train the model in unsupervised manner, we innovate the training procedure by formalizing it as a minimum-cost flow problem. We incorporate prior linguistic knowledge into the design of both model architecture and training in order to  properly guide the decipherment learning process. We conducted experiments on cognate identification tasks for Romance languages, Ugaritic, and for the first time, Linear B. In all cases our system reached high accuracy and significant relative gains over results reported in existing literature.
\end{comment}

\end{abstract}
\section{Introduction}

Decipherment is an ultimate low-resource challenge for both humans and machines. The lack of parallel data and scarce quantities of ancient text complicate the adoption of neural methods that dominate modern machine translation. Even for human experts this translation scenario proved to be onerous: a typical decipherment spans over decades and requires encyclopedic domain knowledge, prohibitive manual effort and sheer luck~\cite{robinson2002lost}. Moreover, techniques applied for the decipherment of one lost language are rarely reusable for another language. As a result, every significant human decipherment is considered to be one of a kind, ``the rarest category of achievement"~\cite{pope1975story}.

Prior work has demonstrated the feasibility of automatic decipherment. \citet{snyder10astatistical} translated the ancient Semitic language Ugaritic into Hebrew. Since both languages are derived from the same proto-Semitic origin, the translation involved matching their alphabets at the character level and mapping cognates at the word level. The effectiveness of their approach stemmed from its ability to incorporate expansive linguistic knowledge, including expected morphological correspondences, the nature of alphabet-level alignment, etc. As with human decipherment, this approach is highly customized for a given language pair and does not generalize to other lost languages. 

In this paper, we introduce a neural decipherment algorithm that delivers strong performances across several languages with distinct linguistic characteristics. As in prior work, our input consists of text in a lost language and a non-parallel corpus in a known related language. The model is evaluated on the accuracy of aligning words from the lost language to their counterparts in the known language. 

To maintain the language-independent nature of the approach, we want to build the model around the most basic decipherment principles applicable across multiple languages. These principles are informed 
by known patterns in language change extensively documented in historical linguistics~\cite{campbell2013historical}. At the character level, we know that characters that originate from the same proto-language have similar distributional profiles with respect to their occurrences. Another important constraint at the character level is that cognate alignment is monotonic since character reorderings within cognate pairs are rare. At the vocabulary level, we want to enforce skewed mapping at the word level assuming roughly one-to-one correspondence. Finally, we want to ensure that the resulting vocabulary mapping covers a significant portion of the lost language vocabulary and can also account for the presence of words which are not cognates. 

Our model captures both character-level and word-level constraints in a single generative framework wherein vocabulary level alignment is a latent variable. We model cognate generation process using a character-level sequence-to-sequence model which is guided towards monotonic rewriting via regularization. Distributional similarity at the character level is achieved via universal character embeddings. We enforce constraints on the vocabulary mapping via minimum-cost flow formulation that controls structural sparsity and coverage on the global cognate assignment. The two components of the model -- sequence-to-sequence character alignment and flow constraints -- are trained jointly using an EM-style procedure.

We evaluate our algorithm on two lost languages -- Ugaritic and Linear B. In the case of Ugaritic, we demonstrate improved performance of cognate identification, yielding 5.5\% absolute improvement over previously published results~\cite{snyder10astatistical}. This is achieved without assuming access to the morphological information in the known language.

To demonstrate the applicability of our model to other linguistic families, we also consider decipherment of Linear B, an ancient script dating back to 1450BC. Linear B exhibits a number of significant differences from Ugaritic, most noted among them its syllabic writing system. It has not been previously deciphered by automatic means. We were able to correctly translate 67.3\% of Linear B cognates into their Greek equivalents in the decipherment scenario.  Finally, we demonstrate that the model achieves superior performance on cognate datasets used in previous work~\cite{kirkpatrick13decipherment}.

\section{Related Work}

\paragraph{Decoding of Ciphered Texts}  Early work on decipherment was primarily focused on man-made ciphers, such as substitution ciphers.  Most of these approaches are based on EM algorithms which are further adjusted for target decipherment scenarios. These adjustments are informed by assumptions about ciphers used to produce the data \cite{knight99acomputational,knight06unsupervised,ravi2011deciphering,pourdamghani2017deciphering}. Besides the commonly used EM algorithm, \cite{nuhn13beam,hauer14solving,kambhatla18decipherment} also tackles substitution decipherment and formulate this problem as a heuristic search procedure, with guidance provided by an external language model (LM) for candidate rescoring. So far, techniques developed for man-made ciphers have not been shown successful in deciphering archaeological data. This can be attributed to the inherent complexity associated with processes behind language evolution of related languages. 

\paragraph{Nonparallel Machine Translation}  Advancements in distributed representations kindled exciting developments in this field, including translations at both the lexical and the sentence level. Lexical translation is primarily formulated as alignment of monolingual embedding spaces into a crosslingual representation using adversarial training~\cite{conneau2017word}, VAE~\cite{dou18unsupervised}, CCA~\cite{haghighi08learning,faruqui14improving} or mutual information~\cite{mukherjee2018learning}. The constructed monolingual embedding spaces are usually of high quality due to the large amount of monolingual data available.  The improved quality of distributed representations has similarly strong impact on non-parallel translation systems that operate at the sentence level~\cite{pourdamghani2017deciphering}.  In that case, access to a powerful language model can partially compensate for the lack of explicit parallel supervision.  Unfortunately, these methods cannot be applied to ancient texts due to the scarcity of available data.

\paragraph{Decoding of Ancient Texts} \cite{snyder10astatistical} were the first to demonstrate the feasibility of automatic decipherment of a dead language using non-parallel data. The success of their approach can be attributed to cleverly designed Bayesian model that structurally incorporated powerful linguistic constraints. This includes customized priors for alphabet matching, incorporation of morphological structure, etc. \cite{kirkpatrick11simple} proposed an alternative decipherment approach based on a relatively simple model paired with sophisticated inference algorithm. While their model performed well in a noise-free scenario — when matching vocabularies only contain cognates, it has not been shown successful in a full decipherment scenario.  Our approach outperforms these models in both scenarios. Moreover, we have demonstrated that the same architecture deciphers two distinct ancient languages — Ugaritic and Linear B. The latter result is particularly important given that Linear B is a syllabic language.

\section{Approach} \label{sec:model_training}

The main challenge of the decipherment task is the lack of strong supervision signal that guides standard machine translation algorithms. Therefore, the proposed architecture has to effectively utilize known patterns in language change to guide the decipherment process. These properties are summarized below:

\begin{enumerate}
\item \emph{Distributional Similarity of Matching Characters}: Since matching characters appear in similar places in corresponding cognates, their contexts should match.
\item \emph{Monotonic Character Mapping within Cognates}: Matching cognates rarely exhibit character reordering, therefore their alignment should be order preserving.
\item \emph{Structural Sparsity of Cognate Mapping}: It is well-documented in historical linguistics that cognate matches are mostly one-to-one, since both words are derived from the same proto-origin.
\item \emph{Significant Cognate Overlap Within Related Languages}: We expect that the derived vocabulary mapping 
will have sufficient coverage for lost language cognates.
\end{enumerate}

\subsection{Generative framework}
We encapsulate these basic decipherment principles into a single generative framework. Specifically, we introduce a latent variable $\mF=\{f_{i,j}\}$ that represents the word-level alignment between the words in the lost language $\mX=\{x_i\}$ and those in the known language $\mY=\{y_j\}$.
More formally, we derive the joint probability
\begin{align}
    \Pr(\mX, \mY) &= \sum_{\mF\in\mathbb{F}}\Pr(\mF)\Pr(\mX|\mF)\Pr(\mY|\mF,\mX) \nonumber \\
                  &\propto \sum_{\mF\in\mathbb{F}}\Pr(\mY|\mX,\mF) \nonumber \\
                  &= \sum_{\mF\in\mathbb{F}}\prod_{y_j\in\mY} \Pr(y_j|\mX,\mF), \label{eq:latent}
\end{align}
by assuming a uniform prior on both $\Pr(\mF)$ and $\Pr(\mX|\mF)$, and i.i.d. for every $y_j\in\mY$. We use $\mathbb{F}$ to describe the set of valid values for the latent variable $\mF$, subject to the global constraints as stated in Property 3 and 4. More specifically, we utilize a minimum-cost flow setup to enforce these properties.

The probability distribution $\Pr(y_j|\mX, \mF)$ is further defined as 
\begin{align}
    \Pr(y_j|\mX, \mF) &= \sum_{x_i\in\mX}  f_{i,j}\cdot\Pr\nolimits_\theta(y_j|x_i), \label{eq:sum_over_x}
\end{align}
where the conditional probability $\Pr\nolimits_\theta(y_j|x_i)$ is modeled by a character-based neural network parameterized by $\theta$, which incorporates the character-level constraints as stated in Property 1 and 2. 

Directly optimizing~\eqref{eq:latent} is infeasible since it contains a summation over all valid flows. To bypass this issue, we adopt an EM-style iterative training regime. Specifically, the training process involves two interleaving steps. First, given the value of the flow $\mF$, the neural model is trained to optimize the likelihood function $\prod_{y_j\in\mY}\Pr(y_j|\mX, \mF)$. Next, the flow is updated by solving a minimum-cost flow problem given the trained neural model. A detailed discussion of the training process is presented in Section~\ref{sec:training}.

We now proceed to provide details on both the neural model and the minimum-flow setup.

\subsection{Neural decipherment model} \label{sec:model}

\begin{figure*}
 \centering
 \includegraphics[width=0.85\linewidth]{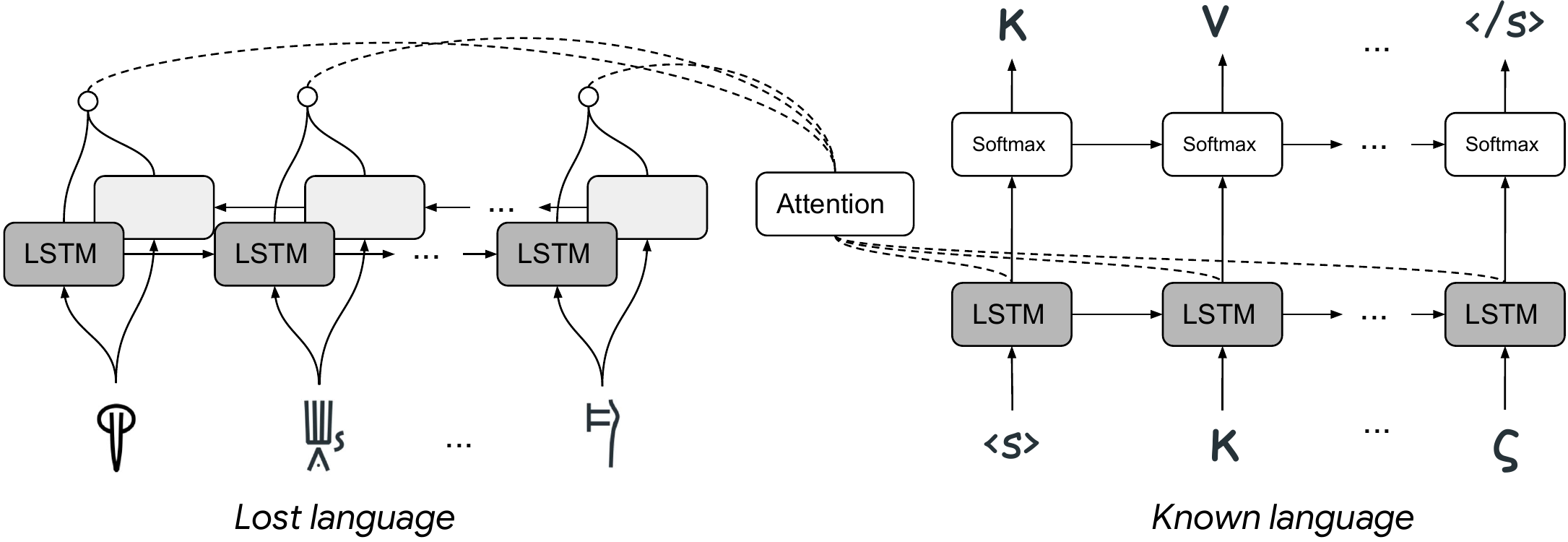}
 \caption{Architecture of our proposed model. For simplicity, we omit lines for residual connections linking weighted sum of input embeddings and softmax. Inputs to the encoder and decoder are the lost and known languages respectively. See Sec.~\ref{sec:model} for details.}
 \label{fig:seq2seq_model}
\end{figure*}

%Our neural decipherment model is based on the attention-based sequence-to-sequence architecture proposed by~\citet{luong2015effective}. 

% To tailor the model towards our task, we make several important enhancements to incorporate the following linguistic principles underlying decipherment:
% \begin{itemize}
%     \item \textbf{Regular Character mapping}: We employ a shared universal character embedding to model the regular correspondences between the characters of two languages. 
%     \item \textbf{Local translation}: We design a norm-controlled residual layer to force the model to mainly consider local character information. 
%     \item \textbf{Monotonic transduction}: We enforce an edit distance based regularization term on the character alignment to guide the model towards monotonic transduction. 
% \end{itemize}
% We detail each component in the following sections.

We use a character-based sequence-to-sequence (seq2seq) model to incorporate the local constraints~(\figref{fig:seq2seq_model}).  
Specifically, we integrate Property 1 by using a shared universal character embedding space and a residual connection. Furthermore, the property of monotonic rewriting is realized by a regularization term based on edit distance. We detail each component in the following paragraphs. 

\paragraph{Universal character embedding}
We directly require that character embeddings of the two languages reside in the same space. Specifically, we assume that any character embedding in a given language is a linear combination of universal embeddings. More formally, we use a universal embedding matrix $U\in M^{n_u \times d}$, a lost language character weight matrix $W_x\in M^{n_x \times n_u}$ and a known language character weight matrix $ W_y\in M^{n_y\times n_u}$. We use $n_u$ to denote the size of the universal character inventory, and $n_x, n_y$ the number of unique characters in the lost and the known languages, respectively. 
Embedding matrices for both languages are  computed by
\begin{align}
E_x &= W_x U, \nonumber \\
E_y &= W_y U. \nonumber 
\end{align}
This formulation reflects the principle underlying crosslingual embeddings such as MUSE~\cite{conneau2017word}. Along a similar line, previous work has demonstrated the effectiveness of using universal word embeddings, in the context of low-resource neural machine translation~\cite{gu2018universal}. 

\paragraph{Residual connection}
Character alignment is mostly local in nature, but this fact is not reflected by how the next character is predicted by the model. Specifically, the prediction is made based on the context vector $\tilde{h}$, which is a nonlinear function of the hidden states of the encoder and the
 decoder. As a result, $\tilde{h}$ captures a much wider context due to the nature of a recurrent neural network. 
 
 To address this issue and directly improve the quality of character alignment, 
we add a residual connection from the encoder embedding layer to the decoder projection layer. Specifically, letting $\alpha$ be the predicted attention weights, we compute
\begin{align}
    c &= \sum_i \alpha_i E_x(i), \nonumber \\
    \hat{h} &= c \oplus \tilde{h}, \label{eq:res}
\end{align}
where $E_x(i)$ is the encoder character embedding at position $i$, and $c$ is the weighted character embedding. $\hat{h}$ is subsequently used to predict the next character. 
A similar strategy has also been adopted by~\citet{nguyen2018improving} to refine the quality of lexical translations in NMT. 

% \todo{mention residual layer somewhere}
% \todo{put this in related work maybe.} Similarly, 

\paragraph{Monotonic alignment regularization}
\begin{figure}
 \centering
 \includegraphics[width=0.65\linewidth]{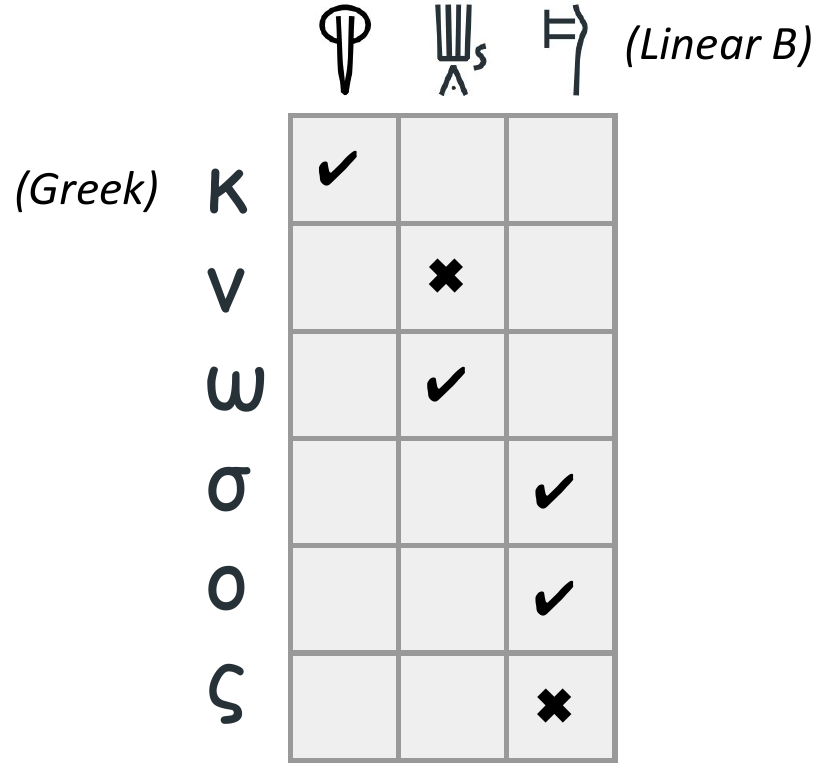}
 \caption{An example of alignment between a Linear B word and Greek word. \ding{52} and \ding{54} denote correct and wrong alignment positions respectively. The misalignment between \textlinb{\Bno} and $\nu$ incurs a deletion error; \textlinb{\Bso} and $\zeta$ incurs an insertion error.}
 \label{fig:align_reg}
\end{figure}

% To encourage the model to learn a mapping that respects linguistic regularity, we impose a regularization term to the attention matrix so that the edit distance between a ciphertext and plaintext is constrained.

We design a regularization term that guides the model towards monotonic rewriting.  Specifically, we penalizes the model whenever insertions or deletions occur. More concretely, for each word in the lost language $x_i$, we first compute the alignment probability $\Pr(a_i^t|x_i)$ over the input sequence at decoder time step $t$, predicted by the attention mechanism. Then we compute the expected alignment position as
\begin{align*}
    p_i^t = \sum_{k} k\cdot\Pr(a_i^t = k | x_i),
\end{align*}
\rv{where $k$ is any potential aligned position.}
The regularization term is subsequently defined as
\begin{align}
 \Omega_1(\{p_i^t\}) &= \sum_{\rv{t}} (p_i^t - p_i^{t-1} - 1)^2. \label{eq:reg1} %\\
%  w_i^t &= \min(l_i - a_i^{t-1}, 1.0),\label{eq:reg} \notag
\end{align}
Note that no loss is incurred when the current alignment position immediately follows the previous position, namely $p_i^t = p_i^{t-1} + 1$. %$\{w_i^t\}$ indicate how  many characters remain to be rewritten, and there will be no penalty for later steps if the model has already transduced the last character. 
%and $\Omega_1(\{a_i^t\})$ is minimized when $\{a_i^t\}$ is diagonal. 
Furthermore, we use a quadratic loss function to discourage expensive multi-character insertions and deletions. 

For Linear B, we modify this regularization term to accommodate the fact that it is a syllabic language and usually one linear B script corresponds to two Greek letters. Particularly, we use the following regularization term for Linear B
\begin{align}
    \Omega_2(\{p_i^t\}) &= \sum_{t=1} (p_i^t - p_i^{t-2} - 1)^2. \label{eq:reg2}
\end{align}
% It can be verified that $\Omega_2$ is minimized when the alignment matrix has a stretched diagonal shape as in Figure~\ref{fig:align_reg}. 
\figref{fig:align_reg} illustrates one alignment matrix from Linear B to Greek. In this example, the Linear B character \textlinb{\Bno} is supposed to be aligned with Greek characters $\nu$ and $\omega$ but only got assigned to $\omega$, hence incurring a deletion error; \textlinb{\Bso} is supposed to be only aligned to $\sigma$ and $o$, but assigned an extra alignment to $\zeta$, incurring an insertion error.

\subsection{Minimum-cost flow} \label{sec:min_flow}
\label{sec:flow}
% Original Draw file: https://docs.google.com/drawings/d/1YxTgq7mUS4BUZ5NJOIF8rUKh-98NQCq2aMELXAe2Fqk/edit
\begin{figure}[t]
\centering
\includegraphics[width=0.5\textwidth]{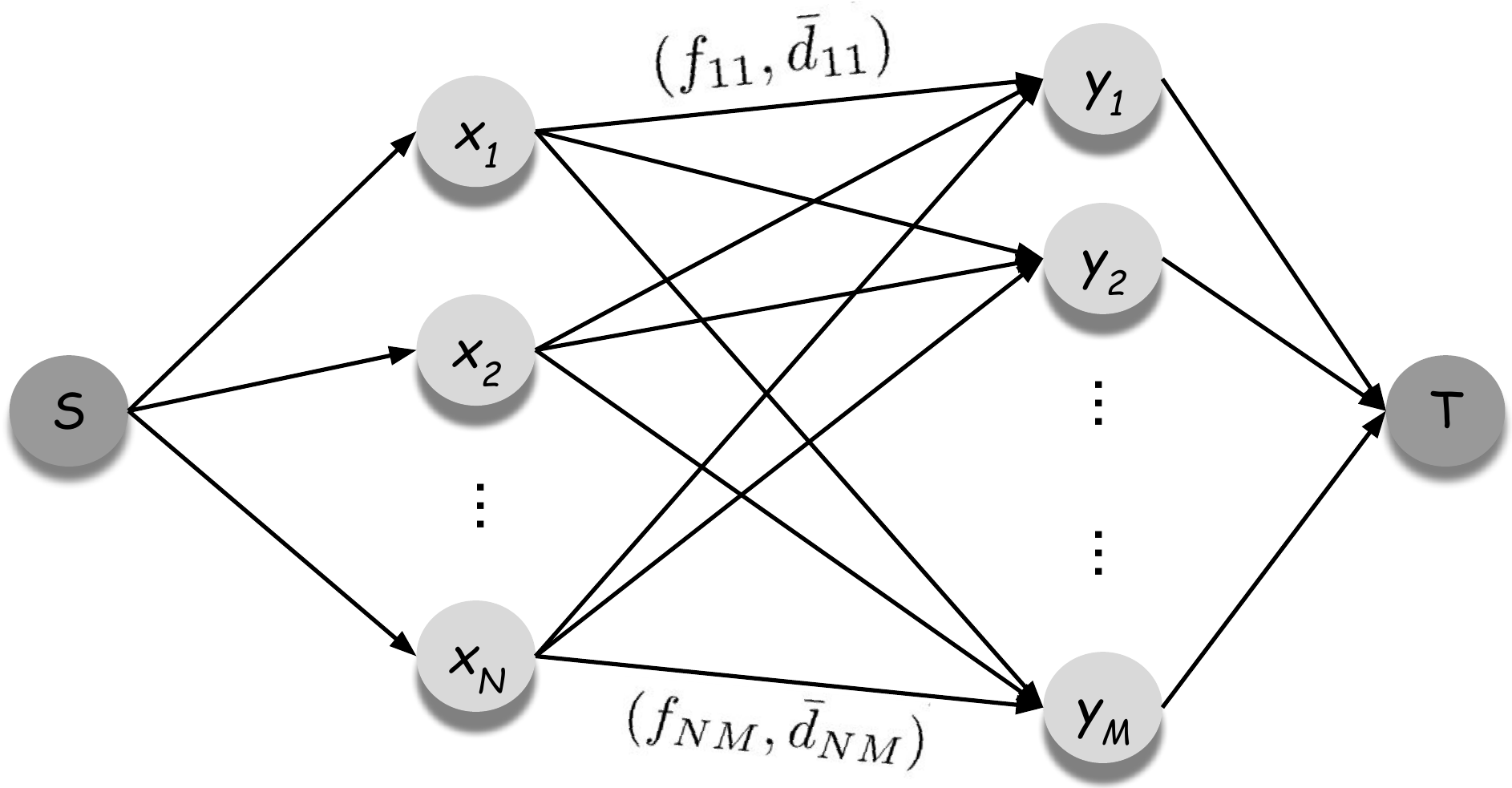}
\caption{Minimum-cost flow. $S$, $T$ stands for source and sink respectively; $x_i$, $y_j$ are the $i^{th}$ and $j^{th}$ word in $\mX$ and $\mY$. Each edge is associated with a flow $f_{ij}$ and cost $\bar{d}_{ij}$. See Sec.~\ref{sec:min_flow} for details.}
\label{fig:flow}
\end{figure}

The latent variable $\mF$ captures the global constraints as stated in Property 3 and 4. 
Specifically, $\mF$ should identify a reasonable number of cognate pairs between the two languages, while meeting the requirement that word-level alignments are one-to-one. To this end, we cast the task of identifying cognate pairs as a minimum-cost flow problem~(\figref{fig:flow}). More concretely, we have three sets of edges in the flow setup:
\begin{itemize}
    \item $f_{s,i}$: edges from the source node to the word $x_i$ in the lost language,
    \item $f_{j,t}$: edges from the word $y_j$ in the known language to the sink node,
    \item $f_{i,j}$: edges from $x_i$ to $y_j$.
\end{itemize}
Each edge has a capacity of 1, effectively enforcing the one-to-one constraint. Only the edges $f_{i,j}$ have associated costs. We define this cost as the expected distance between $x_i$ and $y_j$:
\begin{align}
\bar{d}_{i,j} &= \E\limits_{y \sim \Pr(y|x_i)} d(y, y_j),
\end{align}
where $d(\cdot, \cdot)$ is the edit distance function, and $\Pr(y|x_i)$ is given by the neural decipherment model. We use a sampling procedure proposed by~\citet{shen2016minimum} to compute this expected distance. To provide a reasonable coverage of the cognate pairs, we further specify the demand constraint $\sum_j f_{j,t}= D$ with a given hyperparameter $D$. 

We note that the edit distance cost plays an essential role of complementing the neural model. Specifically, neural seq2seq models are notoriously inadequate at capturing insertions and deletions, contributing to many issues of overgeneration or undergeneration in NMT~\cite{tu2016modeling}. These problems are only accentuated due to a lack of supervision. Using edit distance in the flow setup helps alleviate this issue, since a misstep of insertion or deletion by the neural model will still generate a string that resembles the ground truth in terms of edit distance. In other words, the edit distance based flow can still recover from the mistakes the neural model makes.

\section{Training}\label{sec:training}
% Training under our generative framework is not straightforward, and we employ several techniques detailed below that facilitate efficient and effective training. 

% \subsection{Iterative training}

We note that with weak supervision, a powerful neural model can produce linguistically degenerate solutions. To prevent the neural model from getting stuck at an unreasonable local minimum, we make  three modifications detailed in the following paragraphs. The entire training procedure is illustrated in Alg~\ref{alg:iterative_training}.

\paragraph{Flow decay}
The flow solver returns sparse values -- the flow values for the edges are mostly zero. It is likely that this will discard many true cognate pairs, and the neural model trained on these sparse values can be easily misled and get stuck at some suboptimal local minimum. 

To alleviate this issue, we apply an exponential decay to the flow values, and compute an interpolation between the new flow result and the previous one. Specifically, we update the flow at iteration $\tau$ as
\begin{align}
    f_{i,j}^{(\tau)} &= \gamma\cdot f_{i,j}^{(\tau - 1)} + (1-\gamma)\cdot \tilde{f}_{i,j}^{(\tau)}, \forall i,j, \label{eq:flow_decay}
\end{align}
where $\tilde{f}_{i,j}^{(\tau)}$ is the raw output given by the flow solver, and $\gamma$ is a hyperparameter.

\paragraph{Norm control}
Recall that the residual connection combines a weighted character embedding $c$, and a context vector $\tilde{h}$~(\eqref{eq:res}). We observe that during training, $\tilde{h}$ has a much bigger norm than $c$, essentially defeating the purpose of improving character alignment by using a residual connection. To address this issue, we rescale $\tilde{h}$ so that the norm of $\tilde{h}$ does not exceed a certain percentage of the norm of $c$. More formally, given a ratio $r<1.0$, we compute the residual output as 
\begin{align}
    \hat{h} &= c \oplus (g \cdot \tilde{h}) \nonumber \\
    g &= \min(r * \frac{\norm{c}}{\norm{\tilde{h}}}, 1.0)\nonumber
\end{align}

\paragraph{Periodic reset}
We re-initialize the parameters of the neural model and reset the state of the optimizer after each iteration. Empirically, we found that our neural network can easily converge to a suboptimal local minimum given a poor global word-level alignment. Resetting the model parameters periodically helps with limiting the negative effect caused by such alignments.

\begin{algorithm}[t]
  \caption{Iterative training}
    \label{alg:iterative_training}
  \begin{algorithmic}[1]
    \Require{\Statex $\mX$, $\mY$: vocabularies,
             \Statex $T$: number of iterations,
             \Statex $N$: number of cognate pairs to identify.}
  \State $f_{i,j}^{(0)} \gets\frac{N}{\card{\mX}\cdot\card{\mY}}$ \Comment{Initialize}
  \For{$\tau \gets 1 \text{ to } T$}
    \State $\theta^{(\tau)}$ $\gets$ \textsc{MLE-train}($f_{i,j}^{(\tau-1)}$)
    \Let{$\bar{d}_{i,j}^{(\tau)}$}{$\textsc{Edit-dist}(x_i, y_j, \theta^{(\tau)})$} 
    \Let{$\tilde{f}_{i,j}^{(\tau)}$}{$\textsc{Min-cost-flow}(\bar{d}_{i,j}^{(\tau)})$}
    \Let{${f}_{i,j}^{(\tau)}$}{$\gamma\cdot{f}_{i,j}^{(\tau - 1)} + (1-\gamma)\cdot\tilde{f}_{i,j}^{(\tau)}$}
    \State \textsc{Reset}($\theta^{(\tau)}$)
    
  \EndFor
  \State \Return{$f_{i,j}^{(T)}$}
  
  \Statex
  \Function{\textsc{MLE-train}}{$f_{i,j}^{(\tau)}$}
    \State $\theta^{(\tau)} \gets 
    \argmax_\theta \prod_{y_j\in\mY}\Pr\nolimits_\theta(y_j|\mX, \mF)$
  \State \Return{$\theta^{(\tau)}$}
  \EndFunction

  \end{algorithmic}
\end{algorithm}

% \subsection{Efficient computation}

% Naively computing~\eqref{eq:sum_over_x} requires summing over all possible In supervised learning, seq2seq model training procedure often requires teacher-forcing, that is ground-truth labels are fed to the decoder inputs at each time step. This is not feasible for our problem setting since we do not have labels available, and it would be intractable to consider all possible pairs of $(c_i, p_j)$ as a naive implementation needs to compute $\card{\mC}\times\card{\mP}$ sequence pairs.

% To circumvent this expensive operation while keeping the objective differentiable, we use soft character embeddings as input to the decoder as surrogate teacher-forcing signal. More specifically, the input to the decoder at time step $t+1$ is the expected embedding given by the distribution over characters from step $t$
% \begin{align}
% 	emb_{dec} = \sum_{char\in\mA} Pr(p^t=char|c_i) emb_{tgt}(char)  \nonumber 
% \end{align}

\section{Experiments} \label{sec:experiment}
\paragraph{Datasets}
We evaluate our system on the following datasets:
\begin{itemize}
    \item UGARITIC: Decipherment from Ugaritic to Hebrew. Ugaritic is an ancient Semitic language closely related to Hebrew, which was used for the decipherment of Ugaritic. This dataset has been previously used for decipherment by~\citet{snyder10astatistical}.
    \item Linear B: Decipherment from Linear B to Greek. Linear B is a syllabic writing system used to write Mycenaean Greek dating back to around 1450BC. Decipherment of a syllabic language like Linear B is significantly harder, since it employs a much bigger inventory of symbols (70 in our corpus), and the symbols that have the same consonant or vowel look nothing alike\footnote{For instance, \textlinb{\Bka}, \textlinb{\Bke} and \textlinb{\Bte} encode ``ka", ``ke" and ``te", respectively.}. 
    
    We extracted pairs of Linear B scripts \rv{(i.e., words)} and Greek pronunciations from a compiled list of Linear B lexicon\footnote{\url{https://archive.org/details/LinearBLexicon/page/n5}}. We process the data by removing some uncertain translations, eventually retaining 919 pairs in total. The linear B scripts are kept as it is, and we remove all diacritics in the Greek data. 
    
    We also consider a subset of the Greek data to simulate an actual historical event where many linear B syllabograms were deciphered by being compared with Greek location names. On the Greek side, we retain 455 proper nouns such as locations, names of Gods or Goddesses, and personal names. The entire vocabulary of the Linear B side is kept as it is. This results in a dataset with roughly 50\% unpaired words on the Linear B side. We call this subset Linear B/names.
    
    To the best of our knowledge, our experiment is the first attempt of deciphering Linear B automatically. 
    
    \item ROMANCE: Cognate detection between three Romance languages. It contains phonetic transcriptions of cognates in Italian, Spanish and Portuguese. This dataset has been used by~\citet{hall2010finding} and ~\citet{kirkpatrick11simple}.
    
\end{itemize}
Data statistics are summarized in~\tabref{tab:data_stats}. 

\begin{table*}
    \centering
    \begin{tabular}{lccc}
        \textbf{Dataset} & \textbf{\#Cognates} & \textbf{\#Tokens (lost/known)} & \textbf{\#Symbols (lost/known)} \\
        \toprule
         UGARITIC & 2214 & 7353/41263 & 30/23 \\ 
         Linear B & 919 & 919/919 & 70/28\\
         Linear B/names & 455 & 919/455 & 70/28 \\ 
         ROMANCE & 583 & 583/583 & 25/31/28 (Es/It/Pt)\\
    \end{tabular}
    \caption{Statistics of datasets used in our experiments.}
    \label{tab:data_stats}
\end{table*}

\paragraph{Systems}
We report numbers for the following systems:
\begin{itemize}
    \item \texttt{Bayesian}: the Bayesian model by~\citet{snyder10astatistical} that automatically deciphered Ugaritic to Hebrew
    \item \texttt{Matcher}: the system using combinatorial optimization, proposed by~\citet{kirkpatrick11simple}. 
    \item \texttt{NeuroCipher}: our proposed model.
\end{itemize}
We directly quote numbers from their papers for the UGARITIC and ROMANCE datasets. To facilitate direct comparison, we follow the same data processing procedure as documented in the literature. 

\paragraph{Training details}
Our neural model uses a biredictional-LSTM as the encoder and a single-layer LSTM as the decoder. The dimensionality of character embeddings and the hidden size of LSTM are set to 250 for all experiments. The size of the universal character inventory is 50 for all datasets except Linear B for which we use 100.  The hyperparameter for alignment regularization is set to 0.5, and the ratio $r$ to control the norm of the context vector is set to 0.2. We use \textsc{Adam}~\citep{kingma15adam} to optimize the neural model. To speed up the process of solving the minimum-cost flow problem, we sparsify the flow graph by only considering the top 5 candidates for every $x_i$. $\gamma=0.9$ is used for the flow decay on all datasets except on UGARITIC for which we use $\gamma=0.25$. We use the OR-Tools optimization toolkit\footnote{\url{https://github.com/google/or-tools}} as the flow solver.

We found it beneficial to train our model only on a randomly selected subset (10\%) of the entire corpus with the same percentage of \rv{noncognates}, and test it on the full dataset. It is common for the dataset UGARITIC to contain several cognates for the same Ugaritic word, and we found that relaxing the capacity $f_{j, t}$ to 3 yields a better result. For Linear B, similar to the finding by \citep{kirkpatrick13decipherment}, random restarting and choosing the best model based on the objective produces substantial improvements. In scenarios where many unpaired cognates are present, we follow~\citet{haghighi08learning} to gradually increase the number of cognate pairs to identify.

% In our experiments each iteration starts from optimizing the minimum-cost flow (E-step), followed maximizing the likelihood (M-step) for 20 epochs over the training data. We keep track of mean-accuracy on the training data over a sliding window of 10 epochs, and stop iteration when the mean-accuracy stops increasing. We observed from our experiments that at the beginning of each iteration, randomly perturbing the model and flow parameters helps improve accuracy as they reduce the chance of model getting stuck at the local optima yielded from the previous iteration.\todo{Shall we say randomly re-initialize instead?}

\section{Results}
% We report our results in Table~\ref{table:decipher_results}.

% \begin{table}[]
% \centering
% \begin{tabular}{lccc}
%   \textbf{Model} &\textbf{ROMANCE} &\textbf{UGARITIC} & \textbf{LinearB} \\
%   \toprule
% Bayesian &- &? &- \\
% Matcher &90.1 &? &- \\
% NeuroCipher &91.6 &? &? \\
% \end{tabular}
% \caption{Cognates identification accuracy (\%) on different datasets.}.
% \label{table:decipher_results}
% \end{table}

% We report our system's performance on cognate identification tasks in this section. The metric we use is the accuracy of identified cognates.

\paragraph{UGARITIC}
We evaluate our system in two settings. First, we test the model under the noiseless condition where only cognates pairs are included during training. This is the setting adopted by~\citet{kirkpatrick11simple}. Second, we conduct experiments in the more difficult and realistic scenario where there are unpaired words in both Ugaritic and Hebrew. This is the noisy setting considered by~\citet{snyder10astatistical}. %To be consistent with the experimental settings reported by~\citet{kirkpatrick11simple} and~\citet{snyder10astatistical}, we adopt the former as the baseline for the noiseless condition and the latter for the noisy.  
As summarized by Table~\ref{tab:uga-results}, our system outperforms existing methods by 3.1\% under the noiseless condition, and 5.5\% under the noisy condition. 

We note that the significant improvement under the noisy condition is achieved without assuming access to any morphological information in Hebrew. In costrast, previous system \texttt{Bayesian}
%because in~\citep{snyder10astatistical}, 
utilized an inventory of known morphemes and complete morphological segmentations in Hebrew during training. The significant gains in identifying cognate pairs suggest that our proposed model provide a strong and viable approach towards automatic decipherment. %In contrast, our system makes no such assumption and aim to directly learn correspondence between lexicons. Yet our system outperformed this baseline significantly without taking advantage of prior morphological knowledge about Hebrew. We believe this improvement is mainly due to the minimum-cost flow mechanism, which effectively mitigates the risk of false positive when identifying potential cognate pairs.

\begin{table}[h]
    \centering
    \begin{tabular}{lcc}
    \textbf{System} &\textbf{Noiseless} &\textbf{Noisy} \\
    \toprule
    \texttt{Matcher} & 90.4 & - \\
    \texttt{Bayesian} & - & 60.4 \\
    % Kirkpatrick/Snyder & 90.4 & 60.4\\
    \texttt{NeuroCipher} & \textbf{93.5} & \textbf{65.9} \\
    \end{tabular}
    \caption{Cognate identification accuracy (\%) for UGARITIC under noiseless and noisy conditions. The noiseless baseline result is quoted from~\cite{kirkpatrick11simple}, and the noisy baseline result is quoted from~\cite{snyder10astatistical}.}
    \label{tab:uga-results}
\end{table}

% \todo{characters!!}

\paragraph{Linear B}
 To illustrate the applicability of our system to other linguistic families, we evaluate the model on Linear B and Linear B/names. Table~\ref{tab:linb-results} shows that %. As briefly introduced in Sec.~\ref{sec:linear_b}, deciphering LinearB is a challenging task, more so than UGARITIC due to its syllabic nature, which allows multiple signs corresponding to a common vowel. There is no existing algorithmic decipherment baseline result available, we therefore report numbers provided by our system with no comparison. Similar to UGARITIC, we also consider both noiseless and noisy conditions. From Table~\ref{tab:linb-results}, it can be seen that 
 our system reaches high accuracy at 84.7\% in the noiseless LinearB corpus, and 67.3\% accuracy in the more challenging and realistic LinearB-names dataset. %Despite being a more challenging task, the number for the noisy case is still higher than the UGARITIC case.
 
We note that our system is able to achieve a reasonable level of accuracy with minimal change to the system. The only significant modification is the usage of a slightly different alignment regularization term~(\eqref{eq:reg2}). We also note that this language pair is not directly applicable to both of the previous systems \texttt{Bayesian} and \texttt{Matcher}.
The flexibility of the neural decipherment model is one of the major advantages of our approach. 

% \todo{character mappings}

\begin{table}[t]
    \centering
    \begin{tabular}{lcc}
    \textbf{System} &\textbf{Linear B} &\textbf{Linear B/names} \\
    \toprule
    \texttt{NeuroCipher} & 84.7 & 67.3
    \end{tabular}
    \caption{Cognate identification accuracy (\%) for LinearB under noiseless and noisy conditions.}
    \label{tab:linb-results}
\end{table}

\paragraph{ROMANCE}
Finally, we report results for ROMANCE~\cite{hall2010finding} in Table~\ref{tab:romance}, as further verification of the efficacy of our system. We include the average cognate detection accuracy across all language pairs %(Spanish$\rightleftharpoons$Italian (EsIt), Spanish$\rightleftharpoons$Portuguese (EsPt), Italian$\rightleftharpoons$Portuguese (ItPt)), 
as well as the accuracies for individual pairs. Note that in this experiment the dataset does not contain unpaired words. \tabref{tab:romance} shows that our system improves the overall accuracy by 1.5\%, mostly contributed by Es$\rightleftharpoons$It and It$\rightleftharpoons$Pt.\footnote{We nevertheless observed a slight drop for Es$\rightleftharpoons$Pt. However, for this language pair, the absolute accuracy is already very high ($\geq 95\%$). We therefore suspect that performance on this language pair is close to saturation.}

\begin{table}[h]
    \centering
    \begin{tabular}{lccc|c}
    \textbf{System}  &\textbf{EsIt} & \textbf{EsPt} & \textbf{ItPt} &\textbf{Avg}\\
    \toprule
    \texttt{Matcher} &88.9 &\textbf{95.6} &85.7 & 90.1 \\
    \texttt{NeuroCipher} &\textbf{92.3} &95.0 &\textbf{87.3} & \textbf{91.6}  \\
    \end{tabular}
    \caption{Cognate identification accuracy (\%) for ROMANCE. Avg means the average accuracy across all six language pairs. EsIt, EsPt, ItPt are average accuracy for each language pair respectively (Es=Spanish, It=Italian, Pt=Portuguese). Results for \texttt{Matcher} are quoted from~\cite{kirkpatrick11simple}.}
    \label{tab:romance}
\end{table}

\paragraph{Ablation study}
Finally, we investigate contribution of various components of the model architecture to the decipherment performance. Specifically, we look at the design choices directly informed by patterns in language change:
% (See Sec.~\ref{sec:model})
\begin{table}[]
    \centering
    \begin{tabular}{lr}
    System & UGARITIC \\ 
    \toprule
    \texttt{NeuroCipher}     & \textbf{65.9} \\
    \texttt{\quad -monotonic} & 0.0 \\
    \texttt{\quad -residual} & 0.0 \\
    \texttt{\quad -flow}     & 8.6 
    \end{tabular}
    \caption{Results for the noisy setting of UGARITIC. \texttt{-monotonic} and \texttt{-residual} remove the monotonic alignment regularization and the residual connection, and \texttt{-flow} does not use flow or iterative training.}
    \label{tab:my_label}
\end{table}
In all the above cases, the reduced decipherment model fails.  The first two cases reach  0\% accuracy, and the third one barely reaches 10\%. This illustrates the utmost importance of injecting prior linguistic knowledge into the design of modeling and training, for the success of decipherment. 

% Overall, results provided in this section demonstrated high efficacy of our proposed approach, for languages with both alphabetic and syllabic writing systems. Nevertheless our current system operates purely on character and lexicon level, and we believe performance can be further improved if we take contextual information into account, which can be a natural augmentation of our neural sequence model. We leave this option to future investigation.

% \todo{\subsection{Ablation study}}

\section{Conclusions}
% We proposed a novel approach for decipherment. We utilized a neural sequence model to capture flexible character mapping relationships between ciphertext and plaintext, and innovated a training procedure based on minimum-cost flow. The design of model and training procedure follows our linguistic intuition about decipherment in order to facilitate learning of the system. We conducted experiments on cognate identification tasks for languages of different writing systems, including Ugaritic, Linear B and three Romance languages. Empirical results show state-of-the-art accuracy and substantial improvements over results reported in existing literature. We also demonstrated, for the first time, feasibility of computational decipherment of Linear B.

We proposed a novel neural decipherment approach. We design the model and training procedure following fundamental principles of decipherment from historical linguistics, which effectively guide the decipherment process without supervision signal. We use a neural sequence-to-sequence model to capture character-level cognate generation process, for which the training procedure is formulated as flow to impose vocabulary-level structural sparsity. We evaluate our approach on two lost languages, Ugaritic and Linear B, from different linguistic families, and observed substantially high accuracy in cognate identification. Our approach also demonstrated significant improvement over existing work on Romance languages.
\section*{Acknowledgments}

This research is based upon work supported in part by the Office of the Director of National Intelligence (ODNI), Intelligence Advanced Research Projects Activity (IARPA), via contract \# FA8650-17-C-9116. The views and conclusions contained herein are those of the authors and should not be interpreted as necessarily representing the official policies, either expressed or implied, of ODNI, IARPA, or the U.S. Government. The U.S. Government is authorized to reproduce and distribute reprints for governmental purposes notwithstanding any copyright annotation therein. The authors are also grateful for the support of MIT Quest for Intelligence program. 

\bibliography{acl2019}
\bibliographystyle{acl_natbib}

\end{document}